\documentclass[smallabstract,smallcaptions]{dccpaper}
\usepackage{dsfont, amsmath,amsthm,amssymb,url,mathtools,algorithmic,algorithm}
\usepackage{subcaption, mwe,epsfig,graphicx,color}

\theoremstyle{definition}

\begin{document}
\graphicspath{{figures/}}
\title
{Variational Information Bottleneck on Vector Quantized Autoencoders
}

\author {Hanwei Wu and Markus Flierl \\
School of Electrical Engineering and Computer Science\\
	KTH Royal Institute of Technology, Stockholm\\
	\{hanwei, mflierl\}@kth.se}

\maketitle
\begin{abstract}
In this paper, we provide an information-theoretic interpretation of the Vector Quantized-Variational Autoencoder (VQ-VAE). We show that the loss function of the original VQ-VAE \cite{oord:17:nips} can be derived from the variational deterministic information bottleneck (VDIB) principle \cite{Strouse:18}. On the other hand, the VQ-VAE trained by the Expectation Maximization (EM) algorithm \cite{roy:18} can be viewed as an approximation to the variational information bottleneck(VIB) principle \cite{alemi:17:iclr}.
\end{abstract}
\section{Introduction}	\label{sec:Intro}
The recent advances of variational autoencoder(VAE) provide new unsupervised approaches to learn hidden structure of the data \cite{kingma:14:iclr}. The variational autoencoder is a powerful generative model which allows inference of the learned latent representation. However, the classic VAEs are prone to the \textquotedblleft posterior collapse \textquotedblright phenomenon that the latent representations are ignored due to the powerful decoder. Vector quantized variational autoencoder (VQ-VAE) learns discrete representations by incorporating the idea of vector quantization into the bottleneck stage and the \textquotedblleft posterior collapse \textquotedblright can be avoided \cite{oord:17:nips}. \cite{roy:18} proposes to use the Expectation Maximization algorithm to train the VQ-VAE in the bottleneck stage and achieves higher perplexity of its latent space. Both the proposed VQ-VAE models made progress on training of the discrete latent variable models to match their continous counterparts.

In this paper, we show that the formulation of the VQ-VAEs can be interpreted from an information-theoretic perspective. The loss function of the original VQ-VAE can be derived from the deterministic variational information bottleneck principle \cite{Strouse17}. On the other hand, the VQ-VAE trained by EM algorithm can be viewed as an approximation to the variational information bottleneck. 
\section{Related Work} \label{sec:back}
Given a joint probability distribution $p(\mathbf{X}, Y)$ of input data $\mathbf{X}$ and the observed relevant random variable $Y$, the information bottleneck (IB) method seeks a representation $Z$ such that the mutual information $I(\mathbf{X}, Z)$ is minimized, while preserving the mutual information $I(Z,Y)$ \cite{tishby:99:allerton}. $I(Z,Y)$ can be seen as a measure of the predicative power of $Z$ on $Y$, and $I(\mathbf{X}, Z)$ can be seen as a compression measure. Hence, the information bottleneck is designed to find the trade off between the accuracy and compression. \cite{tishby:15:itw} first used the information bottleneck principle to analysis the deep neural networks theoretically, but no practical models are derived from the IB model. \cite{alemi:17:iclr} presents a variational approximation to the information bottleneck so that the IB-based models can be parameterized by the neural networks.

The deterministic information bottleneck (DIB) principle introduces alternative formulation of the IB problem. It focus on the \emph{representational cost} of the latent $Z$ instead of finding the minimal sufficient statistics for predicting $Y$. Hence, DIB replaces mutual information $I(\mathbf{X}, Z)$ with the entropy $H(Z)$,   Using the similar techniques from \cite{alemi:17:iclr}, \cite{Strouse:18} derived a variational deterministic information bottleneck(VDIB) to approximate the DIB. 

 \section{Variational Information Bottleneck} 
 \label{sec: theory}
 We adapt an unsupervised clustering setting to derive the loss functions of the VDIB and VIB. We denote the data point index $I$ as the input data, the codeword index $Z$ as the latent variable, and the feature representation of input data $\mathbf{X}$ as the observed relevant variable and $\hat{\mathbf{X}}$ as the reconstructed representation. The above variables are subject to the Markov chain constraint
 \begin{equation}
 	\mathbf{X} \leftrightarrow I \leftrightarrow Z\leftrightarrow \hat{\mathbf{X}}.
 \end{equation}
 The information bottleneck principle can be formulated as a rate-distortion like problem \cite{tishby:03}
 \begin{equation}
 \min_{p(Z|I): d_{\text{IB}}(I, Z)\leq D} I(I, Z).
 \end{equation}
 The loss function of the information bottleneck principle is the equivalent problem with the Lagrangian formulation,
 \begin{equation}
 L_{\text{IB}} = d_{\text{IB}}(I, Z) + \beta I(I, Z),
 \end{equation}
 where $\beta$ is the Lagrangian parameter.
 
 Consider the information bottleneck distortion is defined as 
 \begin{equation}
 	d_{\text{IB}}(I, Z) = 	\text{KL}(p(\mathbf{X}|I)\|p(\mathbf{X}|Z)),
 \end{equation}
where $\text{KL}(\cdot)$ denotes the Kullback–Leibler divergence.  Let $\mu$ be the measure on $\mathcal{X}$, and we have $\mathbf{X}\in \mathcal{X}$, $ \hat{\mathbf{X}} \in \mathcal{X}$, we can decompose the $d_{\text{IB}}(I, Z)$ into two terms 
\begin{align}
\text{KL}(p(\mathbf{X}|I)\|p(\mathbf{X}|Z)) &= \sum_{i}\sum_{z} p(i)p(z|i) \int_{\mathcal{X}} p(\mathbf{x}|i)\log \frac{p(\mathbf{x}|i)}{p(\mathbf{x}|z)}d\mu \\
\label{eq: joint}
&= \int_{\mathcal{X}}\sum_{z}p(\mathbf{x}, z)\log \frac{p(z)}{p(\mathbf{x}, z)}d\mu -
\int_{\mathbf{x}} \sum_ip(i, \mathbf{x})\log \frac{p(i)}{ p(i, \mathbf{x})}d \mu,
\end{align}
where (\ref{eq: joint}) is derived from using the chain rule to express the conditional probably $p(\mathbf{x}|z)$ as
\begin{equation}
p(\mathbf{x}|z) = \frac{1}{p(z)}\sum_i p(\mathbf{x}|i)p(z|i)p(i).
\end{equation}
Since the second term of (\ref{eq: joint}) is determined solely by the given data distribution $p(I, X)$ and is a constant, so it can be ignored in the loss function for the propose of minimization. The first term of (\ref{eq: joint}) can have an upper bounded by replacing the $p(\mathbf{x}|z)$ with a variational approximation $q(\hat{\mathbf{x}}|z)$ \cite{alemi:17:iclr}
\begin{align}
\int_{\mathcal{X}}\sum_{z}p(\mathbf{x}, z)\log \frac{p(z)}{p(\mathbf{x}, z)}d \mu  &= -\sum_z p(z)\int_{\mathcal{X}}p(\mathbf{x}|z)\log p(\mathbf{x}|z) d \mu\\
\label{ieq: 1}
& \leq -\sum_z p(z)\int_{\mathcal{X}}p(\mathbf{x}|z)\log q(\hat{\mathbf{x}}|z) d \mu \\
\label{eq:reconstruct}
& = -\int_{\mathcal{X}}\sum_i p(i, \mathbf{x})\sum_{z}p(z|i)\log q(\hat{\mathbf{x}}|z) d \mu,
\end{align}
 where (\ref{ieq: 1}) is resulted from the non-negative of the KL divergence
 \begin{align}
 	\text{KL}(p(\mathbf{X}|Z)\|q(\hat{\mathbf{X}}|Z)) &\geq 0\\
 	\int_{\mathcal{X}} p(\mathbf{x}|z)\log p(\mathbf{x}|z) d \mu &\geq \int_{\mathcal{X}} p(\mathbf{x}|z) \log q(\hat{\mathbf{x}}|z)d \mu.
 \end{align}

Similarly, the mutual information $I(I, Z)$ can have an upper bounded by replacing marginal $p(z)$ with a variational approximation $r(z)$
\begin{align}
I(I, Z) &= \sum_{i}\sum_{z}p(i)p(z|i)\log\frac{p(z|i)}{p(z)} \\
\label{ieq:2}
&\leq \sum_{i}\sum_{z}p(i)p(z|i)\log\frac{p(z|i)}{r(z)} \\
\label{eq:mutual}
&= \text{KL}(p(Z|I) \| r(Z)),
\end{align}
where (\ref{ieq:2}) is resulted from the non-negative of the KL divergence
    \begin{align}
    	\text{KL}(p(Z) \| r(Z)) &\geq 0 \\
        \sum_z p(z)\log p(z) &\geq \sum_z p(z)\log r(z)
    \end{align}
  
    By using the derived upper bound (\ref{eq:reconstruct}) and (\ref{eq:mutual}), we can obtain the loss function of VIB \cite{alemi:17:iclr}
   \begin{equation}
   \label{eq:vib}
   L_{\text{VIB}} = -\int_{\mathcal{X}}\sum_i p(i, \mathbf{x})\sum_{z}p(z|i)\log q(\hat{\mathbf{x}}|z) d \mu  + \beta \text{KL}(p(Z|I) \| r(Z)).
   \end{equation}
   
   The loss function of DIB has the same distortion term $d_\text{IB}$ as the original IB. For the second term, the DIB minimizes the entropy $H(Z)$ of the latent variable instead of the $I(I, Z)$\cite{Strouse17}.  Similarly, the $H(Z)$ can be upper bounded by 
 	  \begin{align}
 	  H(Z) &= -\sum_{z}p(z)\log p(z)  \\
 	  &\leq -\sum_{z}p(z) \log r(z)  \\
 	  &= -\sum_{z}\sum_{i}p(i)p(z|i)\log r(z)  \\
 	  & = H(p(Z|I)), r(Z))
 	  \end{align}
Then, we can obtain the loss function of the VDIB \cite{Strouse:18}
 	   	  \begin{equation}
 	   	  \label{eq:vdib}
 	   	  L_{\text{VDIB}} = -\int_{\mathcal{X}}\sum_i p(i, \mathbf{x})\sum_{z}p(z|i)\log q(\hat{\mathbf{x}}|z) d \mu  + \beta \text{H}(p(Z|I)), r(Z))
 	   	  \end{equation}
 	  \section{Connection to VQ-VAEs}
 	  In this section, we establish the connection between the VIB and VDIB principles with the VQ-VAE and the VQ-VAE trained by EM algorithm. In the VQ-VAE setting, the distribution $p(z|i)$ is parameterized by the encoder neural network $p_{\theta}(\cdot)$ and the distribution $q(\hat{\mathbf{x}}|z)$ is parameterized by the decoder neural network $q_{\phi}(\cdot)$. 
 	  
 	  The loss function of the VQ-VAE uses three terms to minimize the first term of (\ref{eq:vib}) and (\ref{eq:vdib}) empirically \cite{oord:17:nips}
 	\begin{equation}
 	\label{eq:loss1}
 	\begin{split}
 	L_{\text{VQ-VAE}} &= \frac{1}{N}\sum_{i = 1}^{N}\left[-\sum_{z = 1}^{K}p_\theta(z_e(\mathbf{x}_i)|i)\log q_{\phi}(\hat{\mathbf{x}}|z_q(\mathbf{x}_i)) \right.\\
 	&\left.+ \beta\|z_e(\mathbf{x}_i)- sg\left(z_q(\mathbf{x}_i)\right)\|_2^2 + \|\text{sg}\left(z_e(\mathbf{x}_i)\right)-z_q(\mathbf{x}_i)\|_2^2 \right],
 	\end{split}
 	\end{equation}
 	  where $\text{sg}(\cdot)$ is the stop gradient operator, $K$ is the number of codewords of  the quantizer, $z_e(\mathbf{x}_i)$ is the output of the encoder of the $i$-th data point, $z_q(\mathbf{x}_i)$ is the output of the bottleneck quantizer and the input of the decoder. The stop gradient operator outputs its input as it is in the forward pass, and it is not taken into account for computing gradients in the training process.
 	  
 	The first term of (\ref{eq:loss1}) is the reconstruction error between the output and input. The gradients of the backpropagation is copied from the decoder input $z_q(\cdot)$ to the encoder output $z_e(\cdot)$. Hence, the first term only optimizes the encoder and decoder, and the codewords receive no update gradients. The second term is the commitment loss that is used to force the encoder output $z_e(\cdot)$ commits to the codewords and the bottleneck codewords are optimized by the third term. $\beta$ is a constant weight parameter for the commitment loss.
 	  
 	  For the second regularization term, VDIB minimizes the cross entropy with the empirical expression
 	  \begin{equation}
 	  \label{eq:crossentropy}
 	  H(p(Z|I)), r(Z)) = \frac{1}{N}\sum_{i = 1}^{N}\sum_{z = 1}^Kp(z|i)\log r(z) 
 	  \end{equation}
 	 Conventionally, the marginal $r(Z)$ is set to be a uniform distribution. Then $H(p(Z|I)), r(Z))$ becomes a constant and can be omit from the loss function. The loss function of VDIB then can be reduced to the loss function (\ref{eq:loss1}) of VQ-VAE.
 	 
      For the VIB, the KL divergence can be expressed as
 	 \begin{equation}
 	 \label{eq:KLVQ}
 	 	\text{KL}(p(Z|I) \| r(Z)) = H(p(Z|I), r(Z)) - H(p(Z| I))
 	 \end{equation}
 	 The first term of (\ref{eq:KLVQ}) is the cross entropy that the same as (\ref{eq:crossentropy}). However, the conditional entropy $H(p(Z| I))$ of (\ref{eq:KLVQ}) encourages the input data to be quantized uniformly with more codewords. 
 	 
 	 The classic VQ-VAE applies nearest neighbor search on the codebook in the bottleneck stage
 	 \begin{equation}
 	 z = \arg \min_{j \in [K]} \|z_e(x_i)-e_j\|_2, 
 	 \end{equation}
 	 where $e_j, j = 1, \dots, K$ is the codeword. Hence, the conditional entropy $H(p(Z|I)$ is zero. 
 	 
 	 On the other hand, the VQ-VAE trained by the EM algorithm uses a soft clustering scheme based on the distance between the codeword and the output of the encoder. The probability the data assigns with the $z-th$ codeword is
 	 \begin{equation}
 	 p(z|i) = p(z|z_e(\mathbf{x}_i)) = \frac{e^{-\|e_z - z_e(\mathbf{x}_i)\|_2^2}}{\sum_{j = 1}^K e^{-\|e_j- z_e(\mathbf{x})\|_2^2}} 
 	 \end{equation}
 	 That is, the EM algorithm explicitly increases the conditional entropy $H(p(Z|I))$ and achieve a lower value for (\ref{eq:vib}). The experiments in \cite{roy:18} also suggests that VQ-VAE trained by the EM algorithm can achieve higher perplexity of the codewords than the original VQ-VAE. 
 	 \section{Conclusion}
 	 We derive the loss function of VIB and VDIB from a clustering setting. We show the loss function of the original VQ-VAE can be derived from the VDIB principle. In addition, we show that the VQ-VAE trained with the EM algorithm explicitly increases the perplexity of the latents and can be viewed as an approximation of the VIB principle.
\Section{References}
\bibliographystyle{IEEEbib}
\bibliography{fine3}

\begin{thebibliography}{1}

\bibitem{oord:17:nips}
A.~Oord, K.~Kavukcuoglu, and O.~Vinyals,
\newblock ``Neural discrete representation learning,''
\newblock in {\em Advances on Neural Information Processing Systems (NIPS)},
  Long Beach, CA, Dec. 2017.

\bibitem{Strouse:18}
DJ~Strouse and D.~Schwab,
\newblock ``Variational deterministic information bottleneck,'' 2018,
\newblock [Online]. Available: http://djstrouse.com/downloads/vdib.pdf.

\bibitem{roy:18}
A.~Roy, A.~Vaswani, A.~Neelakantan, and N.~Parmar,
\newblock ``Theory and experiments on vector quantized autoencoders,'' 2018,
\newblock [Online]. Available: https://arxiv.org/abs/1803.03382.

\bibitem{alemi:17:iclr}
A.A. Alemi, I.~Fischer, J.~V. Dillon, and K.~Murphy,
\newblock ``Deep variational information bottleneck,''
\newblock in {\em Proceedings of the International Conference on Learning
  Representations (ICLR)}, Toulon, France, Apr. 2017.

\bibitem{kingma:14:iclr}
D.~P. Kingma and M.~Welling,
\newblock ``Auto-encoding variational bayes,''
\newblock in {\em Proceedings of the International Conference on Learning
  Representations (ICLR)}, Banff, Canada, Apr. 2014.

\bibitem{Strouse17}
DJ~Strouse and D.~Schwab,
\newblock ``The deterministic information bottleneck,''
\newblock {\em Neural Comput.}, vol. 29, no. 6, pp. 1611–1630, 2017.

\bibitem{tishby:99:allerton}
N.~Tishby, F.~C. Pereira, and W.~Bialek,
\newblock ``The information bottleneck method,''
\newblock .

\bibitem{tishby:15:itw}
N.~Tishby and N.~Zaslavsky,
\newblock ``Deep learning and the information bottleneck principle,''
\newblock .

\bibitem{tishby:03}
A.~Gilad-Bachrach, A.~Navot, and N.~Tishby,
\newblock ``An information theoretic tradeoff between complexity and
  accuracy,''
\newblock in {\em Proceedings of the COLT}, 2003.

\end{thebibliography}
\end{document}